# Monte-Carlo optimizations for resource allocation problems in stochastic network systems

**Milos Hauskrecht**
Department of Computer Science
5329 Sennott Square
University of Pittsburgh
milos@cs.pitt.edu

**Tomas Singliar**
Department of Computer Science
5802 Sennott Square
University of Pittsburgh
tomas@cs.pitt.edu

## Abstract

Real-world distributed systems and networks are often unreliable and subject to random failures of its components. Such a stochastic behavior affects adversely the complexity of optimization tasks performed routinely upon such systems. In this work we investigate Monte Carlo solutions for a class of two-stage optimization problems in stochastic networks in which the expected value of resources allocated before and after the occurence of stochastic failures needs to be optimized. The limitation of these problems is that their exact solutions are exponential in the number of unreliable network components: thus, exact methods do not scale-up well to large networks often seen in practice. We first show that Monte Carlo optimization methods can overcome the exponential bottleneck of exact methods. Next we support our theoretical findings on resource allocation experiments and show a very good scale-up potential of the methods on problems with large stochastic networks.

## 1 Introduction

Many distributed and network systems in practice are unreliable and subject to random failures of its components. Examples of such systems are power grids, where the distribution ability of the network can be affected by demand overloads, equipment malfunctions, and other random events, or various transportation/communication/information networks subject to congestions and intermittent failures. Stochastic behavior of a network system and its components can affect seriously the functionality of the system and tasks to be performed over its resources and topology.

Many tasks to be accomplished by the network system can be formulated as resource allocation (optimization) problems. In resource allocation problems we seek the best assignment of resources to networks (e.g. flows on edges), typically with some consideration to costs and benefits of such allocations. In the stochastic case, we need to consider random fluctuations of the underlying network structure and associated difficulties in satisfying our objectives. While there are often models and efficient algorithms for allocation problems in deterministic systems, typically formulated as matching or flow optimization problems [18, 6], solutions for the networks with stochastic components are not straightforward. The main difficulty is that the quality of an individual allocation policy depends on all possible configurations of unreliable network components. The problem here is the curse of dimensionality: the number of all possible network outcomes can be exponential in the number of unreliable network components. The challenge is to investigate solutions that can alleviate the exponential bottleneck and can scale-up well to large real-world networks.

The focus of this work is on the development and analysis of Monte Carlo solutions for two-stage resource allocation problems in large stochastic networks. Two-stage stochastic problems were introduced by Dantzig [7]; Birge and Loveaux [2] summarize the advances in this area. From the network-related work, Monte Carlo solutions for two-stage route optimization problems [15] in unreliable networks have been studied recently by Verweij et al [21].

A two-stage problem in context of unreliable networks consists of the following sequence of steps: (1) allocate initial resources over the unreliable network, (2) observe stochastic events (failures of network components), (3) allocate/reallocate the remaining network resources given constraints imposed by initial allocations and network failures. Two allocation stages are subject to different set of constraints. Some allocation decisions can be made only in the first stage before stochastic failures take place; the second stage decisions are constrained by the first step allocations and observed configuration of unreliable network components. Because of the dependencies between allocation stages this type of problem is also referred to as a two-stage problem with recourse [7, 2]. Two-stage stochastic problems arise commonly in various investment, supply management or communication applications. Examples include optimizations of product distribution in supply management subject to transportation failures, trading in distributed commodity markets, optimizations of equipment parameters in communication networks, allocation of ancillary generating capacities in power grids and others.

A two-stage resource allocation problem is a special case of a stochastic planning problem. Stochastic planning problems has been studied extensively by researchers in AI for a number of years. Much of the research work has focused on planning problems with very large state spaces [3, 8, 4, 13, 9], considerably less attention has been paid to problems with complex action spaces [17, 10]. Although two stage resource allocation problems in stochastic networks are restricted in terms of the number of decision stages, their state and action spaces are very complex and include many dependent components. An interesting aspect of the problem that affects tremendously the complexity of the optimization is the random fluctuation of the underlying network topology and the fact that optimizations of resources are performed over such topologies.

In the following text, we first formulate the two-stage stochastic problem for network systems and describe the exponential bottleneck associated with its exact solution. Next we describe Monte Carlo solutions for the evaluation and optimization of the two-stage problems and analyze their complexity. Finally we test the scale-up potential of the algorithms on a problem of optimal allocations of consumer and producer capacities in a stochastic transportation network and illustrate a very good performance of methods in terms of the quality of obtained approximations and their running times.

## 2 Two-stage stochastic optimization problem

The goal of the two-stage problem is to allocate resources over the unreliable stochastic network such that we maximize the expected utility of such allocations.

Let $\mathbf{x} = \{x_1, x_2, \cdots, x_m\}$ denote a vector of allocation decisions made in the first stage; $\mathbf{X}$ a space of all possible allocation choices; $\mathbf{s} = s_1 s_2 \cdots s_k$ a failure configuration – an outcome of a multivariate random variable with binary (0,1) values, such that 0 represents the failure and 1 properly operating component; $\mathbf{S}$ a space of all possible configurations, and, $\mathbf{y}(\mathbf{s})$ a vector of allocations selected in the second stage. The two-stage allocation problem (with recourse) can be formulated as:

maximize$_{(\mathbf{x})}$ $Q(\mathbf{x}) = f_1(\mathbf{x}) + E_\mathbf{s}(f_2(\mathbf{x}, \mathbf{s}))$
subject to   $g_1(\mathbf{x}) \leq 0, \cdots, g_u(\mathbf{x}) \leq 0$
             $h_1(\mathbf{x}, \mathbf{y}(\mathbf{s})), \cdots, h_v(\mathbf{x}, \mathbf{y}(\mathbf{s})) \leq 0 \ \forall \mathbf{s} \in S$

Function $Q(x)$ defines expected value (utility) of the allocation $\mathbf{x}$. It decomposes into the first-stage value $f_1$, and the expectation over all second stage values $f_2$. Note that $f_2$ itself corresponds to a mathematical optimization problem. Functions $g_i$ and $h_i$ define the constraints variables must satisfy. Constraints $h_1, \cdots, h_v$ describe the dependencies between the first stage decisions $\mathbf{x}$ and the second stage decisions $\mathbf{y}(\mathbf{s})$. Each constraint holds with probability 1, or for each possible $\mathbf{s} \in \mathbf{S}$.

A two stage optimization problem (with two optimization stages) is typically converted to a single optimization problem where allocations $\mathbf{x}$ are optimized together with all possible second stage decision $\mathbf{y}(\mathbf{s})$.

**Curse of dimensionality.** The hardest part of solving a two-stage recourse problem is to evaluate the expected values of $f_2$. When random variables are discretely distributed, the problem can be written as a large deterministic problem: the expectations can be rewritten as sums, and each constraint can be duplicated for each realization of random variables. The apparent limitation here is the curse of dimensionality; the complexity of the formulation depends on the space of all realizations of random variables $\mathbf{S}$. For discrete variables, the space is exponential in the number of variables.

We stress that the dimensionality problem affects adversely not only the optimization but also the evaluation task in which we want to compute the value of some fixed allocation $\mathbf{x}$:

$$Q(\mathbf{x}) = f_1(\mathbf{x}) + E_\mathbf{s}(f_2(\mathbf{x}, \mathbf{s})).$$

**Objective.** Our main objective is to alleviate the dimensionality problem. Basically, we want to eliminate the need to consider explicitly all configurations in $\mathbf{S}$ and replace them with a small (polynomial) set of failure configurations. Then, if the set of constraints and a function $f_1$ are linear and $f_2$ is a linear programming problem the exact solution can be obtained efficiently.

**Solutions to dimensionality problem.** In this work we focus on Monte-Carlo (MC) approximations. Alternative approaches include (1) mean-based approximations [19] in which random variables are replaced with their means, and (2) approximations based on state aggregations (clustering) [11] in which the space of failure configuration is partitioned into a "small" set of nonoverlapping groups and every group is replaced by a single group representative. The limitation of both these approaches is that they do not come with any accuracy guarantees and thus represent heuristic approximations.

## 3 Monte-Carlo approximations

The idea behind Monte-Carlo (MC) approximations is to replace expectations over stochastic configurations $E_S(f_2(\mathbf{x}, \mathbf{s}))$ with their sample estimates.

### 3.1 Monte Carlo evaluation

Let $Q(\mathbf{x})$ be the value of the fixed allocation $\mathbf{x}$. The quantity can be approximated using $N$ samples of random variables $\mathbf{s}^1, \mathbf{s}^2, \cdots, \mathbf{s}^N$ as:

$$Q_N(\mathbf{x}) = \frac{1}{N} \sum_{i=1}^{N} \left[ f_1(\mathbf{x}) + f_2(\mathbf{x}, \mathbf{s}^i) \right].$$

It is easy to see that once the sample $\mathbf{s}^1, \mathbf{s}^2, \cdots, \mathbf{s}^N$ is obtained the problem becomes a deterministic problem and its complexity depends only on $N$.

**Sample complexity analysis.** The sample average $Q_N(x)$ is an unbiased estimator of $Q(\mathbf{x})$ and $E(Q_N(\mathbf{x})) = Q(\mathbf{x})$. The probability of making an error larger than $\epsilon$ for a sample of size $N$, $P(|Q_N(\mathbf{x}) - Q(\mathbf{x})| \geq \epsilon)$, can be bounded using standard concentration inequalities (see, for example, McDiarmid [16]). The result is summarized in the following theorem.

**Theorem 1** *Let $Q_N(\mathbf{x})$ be the sample average approximation of $Q(\mathbf{x})$ for the fixed $\mathbf{x}$. Then:*

- $P(|Q_N(x) - Q(x)| \geq \epsilon) \leq 2e^{-\left[\frac{2\epsilon^2 N}{q_{d,\mathbf{x}}^2}\right]}$,

- *The number of samples $N$ to obtain the $\epsilon\delta$ estimate is:*
  $N \geq \frac{q_{d,\mathbf{x}}^2}{2\epsilon^2} \log\left(\frac{2}{\delta}\right)$

*where $q_{d,\mathbf{x}} = q_\mathbf{x}^{\max} - q_\mathbf{x}^{\min}$ is the difference between the maximum and minimum possible values $f_2$ can take given $\mathbf{x}$.*

**Proof.** The quality of the approximation of $Q(\mathbf{x})$ follows directly from Hoeffding's inequality [12]. ∎

The important feature of the approximation is that it is independent of $|S|$: the number of different failure configurations, thus the MC evaluation is able to break the exponential dependency on $|S|$. We note that the bounds in the theorem are not the tightest possible, but they are sufficient to show the independence on $|S|$. Tighter bounds for $N$ can be obtained through Bennet's or Bernstein's concentration inequalities [16] or directly from Chernoff's exponential bound [5].

### 3.2 Monte Carlo optimization

The optimization problem is harder and requires us to optimize over the space $\mathbf{X}$ of all possible first stage allocations. In general, two Monte-Carlo solutions can be applied to solve the problem: interior and exterior methods [19, 14]. In interior methods, sampling is performed inside of a chosen algorithm with new (independent) samples generated in the process. The advantage is that the sampling approach is fitted closely to the optimization method selected. Examples of the methods are: stochastic decomposition and statistical L-shape method. In exterior methods the sample is generated externally, and, thus, it is independent of the optimization method used. In this work we pursue exterior methods.

The idea of the MC optimization is to approximate the allocation problem with the sample average problem [19, 14]:

$$\text{maximize}_{(\mathbf{x})} \quad Q_N(\mathbf{x}) = \frac{1}{N} \sum_{i=1}^{N} \left[ f_1(\mathbf{x}) + f_2(\mathbf{x}, \mathbf{s}^i) \right].$$

The sample $\mathbf{s}^1, \mathbf{s}^2, \cdots, \mathbf{s}^N$ used in the approximation is shared among different allocation choices and the problem is transformed into a deterministic optimization problem. We note that this is possible since allocations of $\mathbf{x}$ do not affect stochastic components and their distribution.

**Sample complexity analysis.** Let $x^* = \arg\max_\mathbf{x} Q(\mathbf{x})$ be the optimal solution to $Q(\mathbf{x})$, and $x_N^* = \arg\max_\mathbf{x} Q_N(\mathbf{x})$ be the optimal solution to the MC optimization $Q_N(\mathbf{x})$. Our goal here is to determine the $\epsilon\delta$ accuracy of the MC solution $x_N^*$. We want to show that similarly to the evaluation problem the precision of the MC method is independent of the size of the configuration space $S$. In the following, we prove this for: (1) finite allocation space $\mathbf{X}$ and (2) allocation space $\mathbf{X}$ bounded to a subspace of $\mathbb{R}^n$. Finite space analysis similar to ours appears in [14]. Complexity results for the bounded subspace are new to our knowledge.

**Theorem 2** *Let $X$ be a finite allocation space of size $|X|$. Then:*

- $P(|Q(x^*) - Q(x_N^*)| \geq \epsilon) \leq 2|X|e^{-\frac{\epsilon^2 N}{2q_d^2}}$.

- *The number of samples that guarantees the $\epsilon\delta$ optimality of $x_N^*$ is: $N \geq \frac{2q_d^2}{\epsilon^2} \log \frac{2|X|}{\delta}$.*

**Proof.** The inequality follows from Hoeffding's bound with Bonferroni's correction. The probability that the estimate of any value $x \in X$ differs by more than $\epsilon/2$ is:

$$P(\exists x \in X \, |Q(x) - Q_N(x)| \geq \epsilon/2) \leq 2|X|e^{-\frac{\epsilon^2 N}{2q_d^2}}.$$

Assuming only differences smaller than $\epsilon/2$ for all $x \in X$, the difference $|Q(x^*) - Q(x_N^*)|$ can be at most $\epsilon$. Thus:

$$P(|Q(x^*) - Q(x_N^*)| \geq \epsilon) \leq 2|X|e^{-\frac{\epsilon^2 N}{2q_d^2}}.$$

The sample complexity bound for $\epsilon\delta$ follows directly by bounding the probability with the confidence parameter $\delta$. ∎

**Bounded allocation space.** Assume, without loss of generality, that $\mathbf{x} \in [0, d]^n$. Let us further assume that the function $Q(\mathbf{x})$ is Lipschitz continuous.

**Theorem 3** *Let $x \in [0,d]^n$ and $Q(x)$ satisfies the Lipschitz continuity condition: $|Q(x) - Q(y)| \leq K_p ||x - y||_\infty$ for any $x, y \in [0,d]^n$. Then:*

- $P(|Q(x^*) - Q(x_N^*)| \geq \epsilon) \leq 2 \left(\frac{2dK_p}{\epsilon}\right)^n e^{-\frac{\epsilon^2 N}{8q_d^2}}.$

- *The number of samples that guarantees the $\epsilon\delta$ optimality of $x_N^*$ is: $N \geq \frac{8q_d^2}{\epsilon^2} \left[ n \log\left(\frac{2dK_p}{\epsilon}\right) + \log\left(\frac{2}{\delta}\right) \right].$*

**Proof.** Let $X_G$ be a finite subset of $X$ such that for all $x \in X$ there exists $x' \in X_G$ satisfying $||x - x'||_\infty \leq \alpha$. $X_G$ can be defined by constructing an n-dimensional equally spaced grid with $\left(\frac{d}{2\alpha}\right)^n$ points.

Let $x_G^*$ be the optimal solution restricted to $X_G$, that is: $x_G^* = \arg\max_{x \in X_G} Q(x)$, and let $x_{N,G}^*$ be the optimal solution to the sample average approximation with $N$ samples restricted to the grid $X_G$, $x_{N,G}^* = \arg\max_{x \in X_G} Q_N(x)$. Then, using the above result for the finite sets the optimal solution satisfies:

$P(|Q(x^*) - Q(x_N^*)| \geq \epsilon) =$

$P(|Q(x^*) - Q(x_G^*) + Q(x_G^*) - Q(x_N^*) + Q(x_{N,G}^*) - Q(x_{N,G}^*)| \geq \epsilon)$

$\leq P(|Q(x_G^*) - Q(x_{N,G}^*)| \geq \epsilon - 2K_p\alpha)$

$\leq 2 \left(\frac{d}{2\alpha}\right)^n e^{-\frac{(\epsilon - 2K_p\alpha)^2 N}{2q_d^2}}$

In the above derivation we used the fact that: $|Q(x^*) - Q(x_G^*)| \leq K_p\alpha$ and $|Q(x_{N,G}^*) - Q(x_N^*)| \leq K_p\alpha$. Substituting $\alpha$ such that $\alpha = \epsilon/4K_p$ we obtain:

$P(|Q(x^*) - Q(x_N^*)| \geq \epsilon) \leq 2 \left(\frac{2dK_p}{\epsilon}\right)^n e^{-\frac{\epsilon^2 N}{8q_d^2}}.$

The number of samples $N$ necessary for $\epsilon\delta$ estimate is obtained by bounding the probability with the confidence parameter $\delta$. ∎

The complexity results in Theorems 2 and 3 show that the number of samples needed to obtain the $\epsilon\delta$ approximation is independent of the size of the configuration space **S** that is exponential in the number of unreliable network components. We see that in the case of the bounded allocation space $[0,d]^n$, the number of samples depends only linearly on the dimension $n$ of the first stage allocation vector. Thus, MC optimization methods eliminate the exponential dependency on the number of unreliable network components.

### 3.3 Practical considerations and improvements

The overall running time performance of the MC optimization depends on two factors: a sample size $N$ and optimization algorithms applied to solve the optimization problem. Since optimization algorithms tend to scale worse than linearly, it is important to keep the number of sample configurations over which we optimize as small as possible. In the following we discuss two solutions addressing this problem.

**Empirical (sample) distribution.** A relatively straightforward simplification reducing the complexity of optimization tasks takes advantage of the fact that configurations can repeat in the sample and thus there are only $N' \leq N$ different configurations to be considered. This effectively reduces the complexity of the MC optimization problem since $Q_N$ can be rewritten using an empirical distribution.

**Subselections from multiple candidate solutions.** The MC optimization can be computationally very expensive since both the first and the second stage decisions must be often optimized in parallel. From this perspective, the MC evaluation of a fixed allocation x with $N$ samples comes with an advantage; the evaluation can be decomposed into $N$ independent second-stage optimization problems of smaller complexity. We can use this property to construct a new class of optimization algorithms. The main idea is that instead of solving one large MC optimization problem defined by a large sample size, it may be computationally less expensive to solve $K$ MC optimization problems based on a smaller sample size $N_1$ and pick the best-out-of $K$ solution by evaluating and comparing each candidate on a larger sample $N_2$.

To illustrate the subselection idea assume that both stages of the two-stage problem correspond to linear problems, and $O(C(u))$ is the complexity of the linear optimization algorithm for the problem of size $u$. Then Monte Carlo evaluation (after the decomposition) can be performed roughly in time $O(N_2 C(m))$ where $N_2$ is the sample size used in the evaluation and $m$ is the complexity of the network. In contrast to this, the complexity of the MC optimization is approximately $O(C(N_1 m))$ where $N_1$ is the sample size for the optimization purposes. Since $O(C(u))$ grows more than linearly, it may be computationally less expensive to first solve $K$ different MC optimization tasks with a smaller sample $N_1$ and then pick the best solution via MC evaluations on a sample size $N_2 > N_1$ rather than performing the MC optimization on a large sample size directly.

## 4 Evaluation

Sample complexity bounds derived in the previous section, though useful from the theoretical perspective, are often of limited practical importance. The reason is that the sample size $N$ required to achieve the desired accuracy is typically much smaller. To obtain a realistic picture about the performance of the MC optimization methods we conduct a series of experiments on resource optimization problems in stochastic networks.

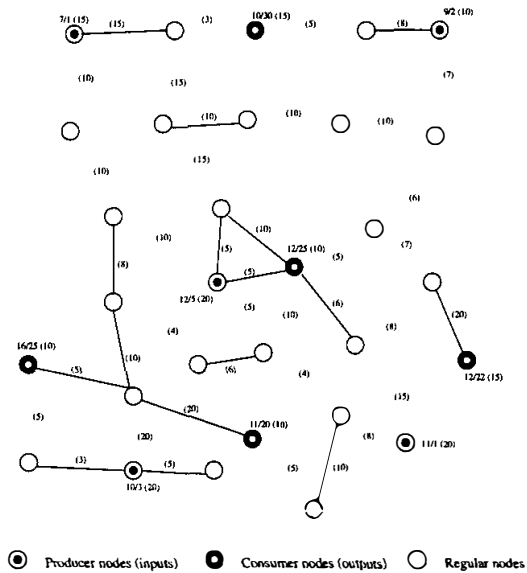

Figure 1: An example of a transportation network with five producer and five consumer nodes. Dashed lines represent unreliable connections subject to random failures. Capacity limits on edges and nodes are shown in parenthesis. Producer and consumer nodes have two prices: purchase (first-stage) and return (second stage) price.

### 4.1 Example

As a example we consider the problem of allocation of producers' and consumers' capacities over an unreliable transportation network. Figure 1 illustrates the topology of one such network. The network consists of three types of nodes: producer (or input) nodes, consumer (or output) nodes and regular nodes. Producer and consumer nodes represent producers/consumers of a resource (commodity, service etc.). Links represent transportation connections. Links and nodes have capacities representing transportation limits and limits on the production/consumption.

In the first stage of the problem the amounts of resources to be produced and consumed by individual producers and consumers are chosen. Resources are bought and sold using purchase prices $r_i^{I,in}$ and $r_j^{I,out}$. The initial purchases and sells are performed before resources are distributed via the transportation network. Thus, initial allocations are best viewed as future contracts on the delivery. The underlying transportation network is subject to random connection failures. Whenever a connection fails the delivery of a resource may not be possible. Such a situation may result in losses. Losses are defined in terms of the return (second-stage) prices: $r_i^{II,in}$ and $r_i^{II,out}$. These are prices one would receive or have to pay if the initial contracts are not satisfied. Thus, in the second stage the decision about the distribution of resources leading to minimal losses is sought.

Our ultimate goal is to find the allocation of resources to producers' and consumers' nodes so that the overall expected returns are optimized. The expectation covers combined returns from both stages.

Let x be the vector of initial resource allocations in producers' and consumers' nodes. The two-stage optimization problem can be expressed as:

$$\text{maximize}_\mathbf{x} \quad Q(\mathbf{x}) = f_1(\mathbf{x}) + E_s(f_2(\mathbf{x}, \mathbf{s}))$$

subject to:

$$0 \leq x_i^{out} \leq c_i^{out} \text{ for every consumer node } i;$$
$$0 \leq x_j^{in} \leq c_j^{in} \text{ for every producer node } j$$

where

$$f_1(\mathbf{x}) = \left[ \sum_i r_i^{I,out} x_i^{out} - \sum_j r_j^{I,in} x_j^{in} \right],$$

and $c_i^{out}, c_j^{in}$ are capacity limits on nodes. The function $f_2$ corresponds to the second stage (linear) optimization problem for the failure configuration s:

$$f_2(\mathbf{x}, \mathbf{s}) = \max_{\mathbf{y}^s} f_3(\mathbf{x}, \mathbf{y}^s, \mathbf{s})$$

where:

$$f_3(\mathbf{x}, \mathbf{y}^s, \mathbf{s}) = \sum_j (x_j^{in} - y_j^{in,s}) r_j^{II,in} - \sum_i (x_i^{out} - y_i^{out,s}) r_i^{II,out},$$

and $\mathbf{y}^s$ represent second stage resource allocations for the failure configuration s. The constraints in the second-stage linear program (not shown) assure the satisfaction of capacity constraints and the conservation of flow after component failures.

The two-stage optimization problem can be merged into one (huge) linear program where the optimization of the two stages is performed in parallel:

$$\max_{\mathbf{x}, \mathbf{y}^{s_1}, \mathbf{y}^{s_2}, \cdots, \mathbf{y}^{s_{|S|}}} f_1(\mathbf{x}) + \sum_{s \in S} p(s) f_3(\mathbf{x}, \mathbf{y}^s, \mathbf{s})$$

subject to constraints for the first stage and the union of constraints for every second stage problem. It is important to stress that the combined linear program offers a tremendous amount of structure so optimization techniques based on L-shape cuts (or Benders decompositions) [20] can be utilized to solve it more efficiently.

### 4.2 Experiments

We have evaluated the MC optimization algorithms on multiple transportation networks. To illustrate the scale-up behavior of the MC algorithm we use the example from Figure 1 with 22 unreliable edges and 600 possible network failure configurations.[1] The exact optimization problem for this network can be formulated as an LP with over 63,600 variables. This is still in the reach of our optimization package (*lp_solve*) [1] so we are able to compare the MC solution to the exact solution. The exact solution took $\sim 1000$ seconds.

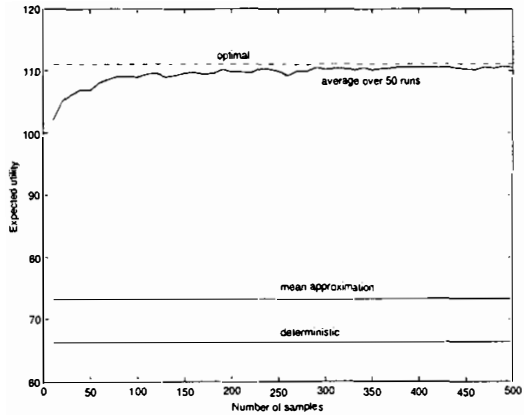

Figure 2: Expected utilities for allocation policies obtained through exact, Monte Carlo, mean-based, and deterministic approximations. Values for MC optimizations are averaged over 50 different solutions.

In the first set of experiments we have compared the MC optimization method for different sample sizes, against the optimal solution and two strawman approximations: the deterministic approximation and the mean-based approximation. The deterministic approximation was obtained by solving the allocation problem while ignoring the failures, the mean-based approximation replaced all random variables representing the edge failures with their means. Figure 2 shows true expected values of allocations for different optimization methods. Because of the variations due to sampling we plot average expected value over 50 different runs of the MC method.

Figure 3 illustrates an average time in seconds (over 50 trials) to solve an MC optimization problem for different sample sizes. In contrast to these results, it took $\sim 1000$ seconds to solve the problem exactly. Solutions of the deterministic and mean-based approximations were obtained in about twenty milliseconds. Fluctuations in average time can be explained by the fact that the MC optimization is performed on the empirical distribution (see section 3.3) that effectively reduces the size of the 'sample' used by the linear program. Despite this, the running times of the MC optimization appear to ramp up faster than linearly in $N$,

---

[1] The total number of failure configurations for the problem in Figure 1 is $2^{22}$. Our initial goal, however, was to formulate a problem we were able to solve exactly so that we can compare the MC solution to the exact solution. For that reason the number of configurations was restricted to 600. Experiments with the full configuration space are described later in the section.

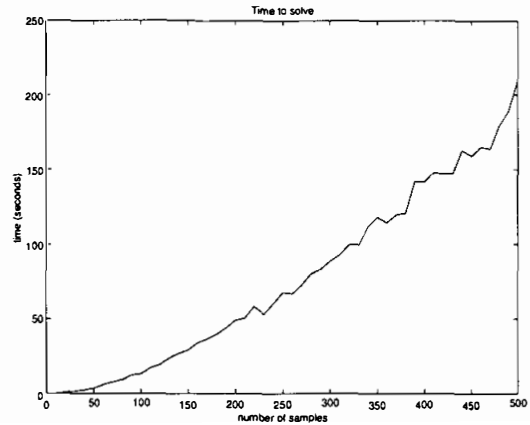

Figure 3: Average time to solve the MC optimization problem for different sample sizes.

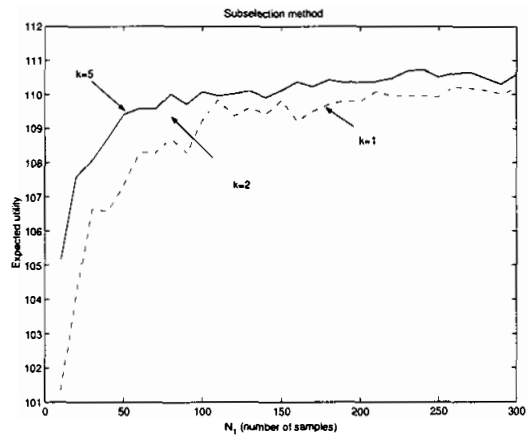

Figure 4: Subselection method. Expected utility obtained for 1, 2 and 5 candidate solutions. The results are averaged over 50 solutions.

which is expected and due to the complexity of the employed LP optimization algorithm.

One way to reduce the complexity of the MC optimization for very complex networks is to use the subselection method (Section 3.3). The approach selects the best allocation from among a set of candidate allocations, each obtained by solving the problem on a smaller sample size $N_1$. The best solution is selected by evaluating and comparing each allocation on a large sample $N_2 > N_1$. The expected benefit is that evaluations are much easier to do since they decompose to a sequence of smaller optimization problems. Figure 4 illustrates the improvement in the quality of allocations found through the subselection method with 2 and 5 candidate solutions as compared to the standard MC optimization ($k = 1$). The number of samples shown in the graph define the sample size $N_1$. The evaluation sample for the subselection method was fixed to $N_2 = 2N_1$. The results in Figure 4 show some improvements in the quality of allocations found through the subselection approach. However, only improvements between the standard MC optimization method ($k = 1$) and the subselection method with $k = 2$ candidates appear to be significant. There is a visible sample gap between two solutions with the same expected utility which indicates improvements in the quality of the solution due to subselection and its potential computational benefits. We note that the sample gap is smaller for smaller sample sizes (in the range of 20-30 samples) and wider for larger sample sizes (50-70 samples). The differences appears to be problem specific but the fact we see a clear separation between the two methods illustrates a potential benefit of the subselection approach and warrants its further study. The differences between subselections with 2 and 5 candidate solutions are less dramatic. This indicates a multiway trade-off between the sample size, the number of candidates used, and the quality of the solution. We expect the optimal balance between these parameters to be problem specific and insights into these trade-offs would require further experimentation.

Overall, the quality of MC optimizations and their running times on our example are very encouraging. The good news is that we have observed similar behaviors on multiple networks of comparable complexity, but with different failure distributions, nodes connectivity and reward structure. To illustrate the scale-up performance of the MC methods on even larger problems we have experimented with a network with 76,000 distinct failure configurations. The exact optimization solution is out of the reach of the current LP optimization methods (the LP would require roughly 8 million variables). However, even in this case we were still able to evaluate the fixed allocations exactly. [2] Figure 5 illustrates the quality of allocations found by the MC methods. The graph shows the average (over 50 runs) of the expected utility of solutions found by the MC optimization method for different sample sizes, and both the minimum and the maximum expected utilities. Although we were not able to compute the optimal solution, the max statistic for 50 trials appears to flatten after about 50 samples and the max value can be considered a reasonable surrogate of the optimal solution.

## 5 Conclusions

We have implemented and studied Monte Carlo optimization and evaluation algorithms for two-stage resource allocation problems in stochastic networks. The limitation of this class of problems is that their complexity grows exponentially in the number of unreliable network components. We showed that Monte Carlo optimization meth-ods are able to eliminate the exponential bottleneck. Since, the MC methods are independent of the space of all failure configurations they have a great potential of scaling-up to very complex real-world network problems. This point has been supported by our experiments on resource allocation problems in transportation networks. These results are also in concordance with very recent experimental results by Verweij et al [21] reporting on the performance of MC optimization methods on stochastic routing problems with up to $2^{1694}$ scenarios.

---

[2] As discussed in Section 3.3 the evaluation of the first stage allocation policy can be performed by solving a sequence of optimization tasks of smaller complexity.

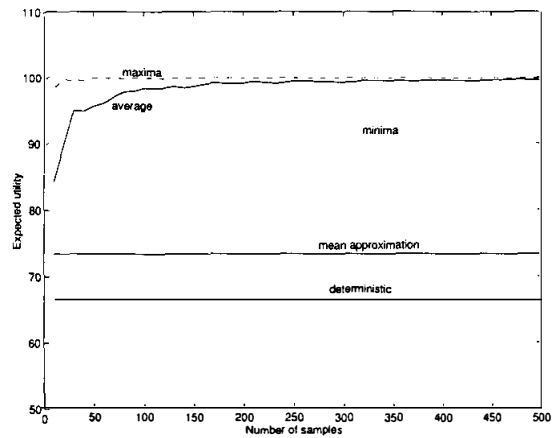

Figure 5: Expected utilities of allocation policies found through the MC optimization method. A network with $\sim 76,000$ different failure configurations was used in the experiment. The results are averaged over 50 solutions. In addition to averages, we also plot minimum and maximum expected values out of 50 averaged policies.

Many interesting research issues related to two-stage allocation problems in stochastic networks remain open. The multiple-candidate method with the subselection from among a smaller set approximate candidate solutions, especially its variance reduction potential, needs to be analyzed in more depth both theoretically and experimentally. A related problem is the choice of the number of candidate solutions to be used by the method. Another challenge is the investigation of the structural properties of real-world network systems in conjunction with multivariate models of its random components. The insights from the analysis can be used to enhance our ability to decompose the problem into a smaller set of optimization subproblems. Finally, two-stage resource allocation problems discussed in this paper are restricted in terms of temporal behaviors they can capture and effects of actions on distributions. The challenge ahead of us is to perform complex resource allocations over unreliable dynamically evolving networks.